# Target Detection of Safety Protective Gear Using the Improved YOLOv5

Hao Liu
*College of Big Data and Information Engineering*
*Guizhou University*
Guiyang, China
ie.hliu21@gzu.edu.cn

Xue Qin*
*College of Big Data and Information Engineering*
*Guizhou University*
Guiyang, China
xqin@gzu.edu.cn
*Corresponding author

***Abstract*—In high-risk railway construction, personal protective equipment monitoring is critical but challenging due to small and frequently obstructed targets. We propose YOLO-EA, an innovative model that enhances safety measure detection by integrating ECA into its backbone's convolutional layers, improving discernment of minuscule objects like hardhats. YOLO-EA further refines target recognition under occlusion by replacing GIoU with EIoU loss. YOLO-EA's effectiveness was empirically substantiated using a dataset derived from real-world railway construction site surveillance footage. It outperforms YOLOv5, achieving 98.9% precision and 94.7% recall, up 2.5% and 0.5% respectively, while maintaining real-time performance at 70.774 fps. This highly efficient and precise YOLO-EA holds great promise for practical application in intricate construction scenarios, enforcing stringent safety compliance during complex railway construction projects.**

## I. Introduction

TIn recent years, there has been significant policy and economic support for infrastructure development worldwide, which has presented both opportunities and challenges for the railway industry[1-3]. The construction industry is characterized by inherent high-risk factors, leading to frequent occurrences of accidents on a global scale. A comprehensive analysis of work-related fatal injuries in Great Britain during the 2021/22 fiscal year, as reported by the UK Health and Safety Executive (HSE), revealed a total of 123 fatalities. Notably, the construction sector accounted for a significant portion of these tragic incidents, with 30 fatalities, representing 24.3% of the overall figure. Disturbingly, 18 fatalities were attributed to accidents involving moving objects[4]. The construction industry stands out as a sector with a significant number of accidents, as highlighted in the 2022 Global Safety Report[5] published by IPAF. According to the report, the construction industry accounted for a staggering 43% of all reported accidents. These distressing statistics underline the critical necessity for the implementation of more stringent regulatory measures aimed at mitigating accidents within the industry. Therefore, in railway construction, accurate supervision of workers wearing safety protective gear is of paramount importance. By monitoring the proper usage of hardhats, reflective vests, and other protective gear, it ensures the personal safety of construction personnel and helps prevent safety accidents from occurring. Currently, the supervision of wearing protective gear primarily relies on manual methods, which not only consumes human resources but is also prone to errors.

Aarti Bansal proposed the use of Radio Frequency Identification (RFID) tags to monitor worker safety[6], while Ana Arboleya suggested an RFID-based real-time tracking system to prevent safety accidents[7]. However, this approach also has its limitations. RFID tag design has some drawbacks, such as limited range, susceptibility to interference, line-of-sight requirements, challenges with tag orientation and placement, environmental factors, power requirements, data security and privacy concerns, as well as cost considerations.

With the advancement of computer vision technology, the use of object detection techniques to analyze construction site surveillance videos enables the automatic detection of the condition of wearing protective gear. With the advancement of computer vision technology, the use of object detection techniques to analyze construction site surveillance videos enables the automatic detection of the condition of wearing protective gear. This approach can achieve accurate and real-time supervision of construction site safety, eliminating the need for manual monitoring and reducing the potential for errors. Doungmala et al. proposed the integration of Haar-like features and circle Hough transform for hardhats detection[8]. Yan et al., on the other hand, employed depth-wise separable convolution in Darknet-53[9] for hardhats detection[10]. F. Zhou et al. use YOLOv5 model with different parameters for hardhats detection[11]. While the aforementioned research and applications have demonstrated the feasibility of using computer vision technology for construction site safety detection, they primarily focus on hardhats detection and do not consider the collaborative detection of multiple types of safety protective gear. In railway construction sites, the category of protective gear includes various types such as hardhats, reflective vests, and more. On the other hand, due to the complexity of the railway construction site environment, there are still additional challenges in the detection of wearing safety protective gear using object detection techniques. Firstly, the placement of monitoring equipment is often distant from the construction site, resulting in a small proportion of hardhats and reflective vests being captured in the overall image. its effective features can be extracted less, thus making it more difficult to identify protective equipment. Secondly, real-time monitoring is another crucial factor that needs to be considered in the detection of safety protective gear at construction sites. It is important to ensure timely actions and responses to any potential safety risks or non-compliance with wearing the protective equipment. Furthermore, in railway construction sites, there is a high presence of large equipment and materials, as well as frequent movement of personnel. This results in a significant amount of occlusion, making it challenging to detect safety gear accurately. In summary, in the context of railway construction scenes, utilizing object detection techniques for hardhats detection faces challenges such as small object recognition, occlusion, and real-time requirements.

Currently, object detection algorithms can be broadly categorized into two main types: single-stage object detection algorithms and two-stage object detection algorithms. Two-stage object detection algorithms generate prior bounding boxes first and then perform object detection and classification. Representative algorithms in this category include SPP-net[12], R-CNN[13], Fast R-CNN[14], Faster R-CNN[14] and others. Although these algorithms achieve high accuracy, they often suffer from a large number of model parameters and slower detection speeds, making it challenging to achieve real-time detection. On the other hand, single-stage object detection algorithms directly extract features from the network and predict the position and category of objects. These algorithms have faster processing speeds, strong generalization capabilities, and are more suitable for detection tasks in construction sites. Typical algorithms in the single-stage object detection category include SSD[15] and YOLO[16]. Among them, SSD achieves a relatively balanced trade-off between accuracy and speed. However, even with this balance, it may still fall short of meeting the requirements for real-time detection. Ultralytics introduced the YOLOv5 algorithm based on YOLO[16], which has fewer parameters and lower complexity. However, it has lower accuracy in detecting small objects, and occasional false positives can occur. This issue significantly affects the detection of safety gear for workers in railway construction sites. To address the issue of low accuracy in detecting small objects, Jin et al. proposed integrating the DWCA mechanism into the backbone network[17]. Similarly, Yan et al. proposed improving the accuracy of hardhats detection by utilizing a multi-scale feature fusion structure to capture shallower information[18]. While these methods can improve the accuracy of detecting small objects and enhance the detection of occluded targets, they also introduce additional model parameters and computational complexity, leading to a decrease in the network's inference speed. As a result, real-time detection cannot be achieved with these approaches. Therefore, the challenges related to small object size and insufficient feature extraction caused by target occlusion still remain to be addressed.

In light of the challenges posed by the small size of objects and partial occlusion in railway construction site environments, which make object detection and feature extraction difficult, this study focuses on the detection of safety gear usage among construction workers at railway sites. This paper proposes the integration of the ECA mechanism into the backbone network of the popular object detection algorithm YOLOv5. In comparison to other attention mechanisms, this method introduces a minimal increase in the number of parameters. It employs a non-dimensional reduction local cross-channel interaction strategy, coupled with an adaptive selection of one-dimensional convolution kernel size, to precisely determine the coverage of local cross-channel interaction. Consequently, it facilitates a more targeted extraction of valuable image feature information, effectively mitigating the difficulties associated with detecting small objects. Due to the frequent occurrence of object occlusion, the presence of containment and being-contained relationships between predicted and ground truth bounding boxes during training can pose challenges. The original GIoU loss function remains unchanged in such scenarios, making it difficult to determine the positional relationship between predicted and ground truth bounding boxes accurately. To address this issue, the EIoU function is introduced as a replacement for the original GIoU function. This modification enables the model to achieve more precise fitting in datasets with significant object occlusion. Combining these approaches, we propose a high-accuracy and fast inference speed neural network architecture called YOLO-EA. The proposed method demonstrates excellent performance on a dataset consisting of practical railway construction site images. YOLO-EA outperforms the YOLOv5 model in terms of both detection accuracy and inference speed for protective equipment detection. This makes YOLO-EA well-suited for applications in natural environments found in railway construction sites.

## II. Material and Methods

The YOLO-EA network model proposed in this paper, as shown in Figure 1, is an enhancement of YOLOv5. By incorporating the ECA attention module into the backbone network of YOLOv5 and replacing the GIoU loss function in YOLOv5's BboxLoss with the EIoU function, this model addresses the challenges of small object detection and occlusion in construction site scenarios.

### *A. The YOLO-EA network structure*

The YOLO-EA network structure proposed in this paper consists of four main components: the input end, backbone network, neck network, and output end, as shown in Figure 1. The input end receives the raw image data. The backbone network is composed of Conv, C3, and SPPF modules, where the SPPF module operates similarly to the spatial pyramid pooling (SPP) by using two small convolutional kernels to extract image features before and after. The ECA attention module, introduced in this paper, is integrated into the backbone network after the SPPF module to enhance the recognition and detection capability of small objects. The neck network includes a feature pyramid and path aggregation structure, which effectively aggregates semantic feature information of different dimensions and propagates bottom-level features to the top. The output end consists of the network prediction module, which predicts targets using multi-scale feature mapping and produces the final results by anchoring the predicted bounding boxes using NMS. The entire network structure utilizes three different scales of output tensors to achieve the detection of safety protective gear wearing.

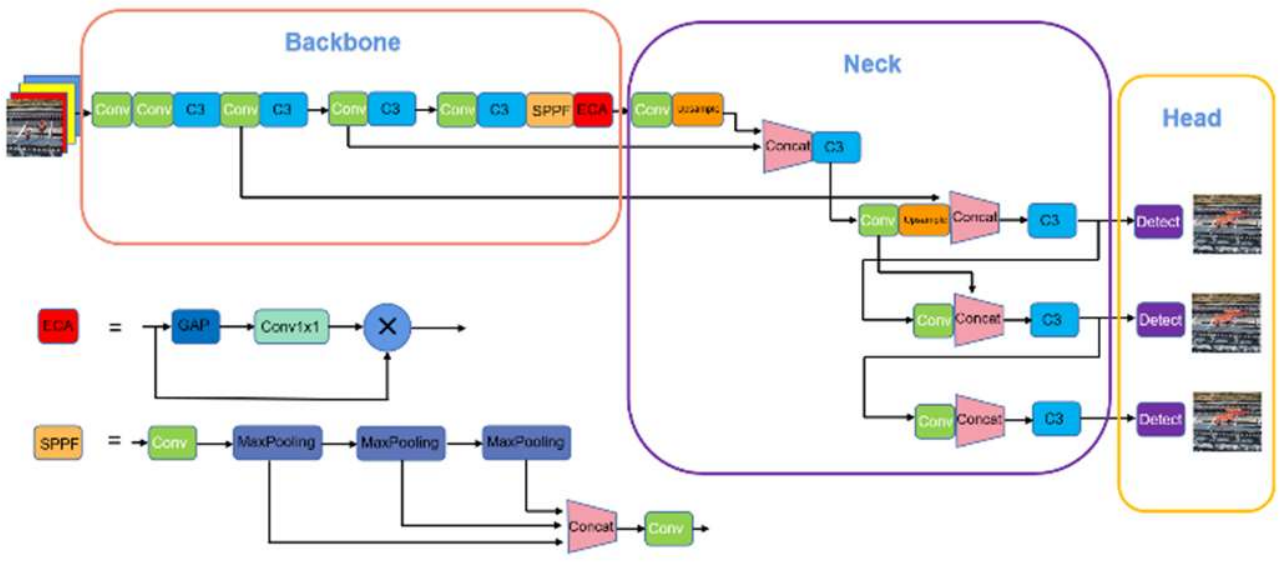

Fig.1 Network structure diagram of YOLO-EA

### B. *ECA attention mechanism module*

In the operational scenarios of railway construction sites, protective equipment often occupies a small portion of the image and has indistinct boundaries. To enhance the network's ability to recognize small objects, an attention mechanism is introduced into the YOLOv5 network model in this paper. In the recognition of protective equipment, although the shape and color features of the protective equipment are crucial information to focus on, the monitoring devices are often positioned at a distance from the actual working scene, resulting in the small objects of protective equipment occupying a very small proportion of the entire image with unclear boundary features. If the model only focuses on easily confusable features, it may lead to missed detections. Therefore, during the training process, the model may be biased towards larger objects, neglecting the feature extraction of small objects. To address this issue, this study incorporates the ECA attention mechanism module[20]. The ECA attention mechanism module helps to extract features from small objects more effectively, thus improving the model's attention to these small objects.

Attention mechanism modules, from the SE attention mechanism[21] to the latest Biformer attention mechanism[22], have significant implications for improving network robustness and accuracy. In fact, numerous studies have demonstrated the effectiveness of attention mechanisms in enhancing model performance, particularly in tasks involving small objects. For instance, SENet is a classic example where the introduction of the SE attention mechanism module significantly improved model performance. Wang et al. proposed the enhancement of small object detection by incorporating the BiCAM attention mechanism module[23]. Building upon these previous research findings, this study introduces the ECA mechanism into the model, enhancing its capability to handle small objects. The structure of the ECA attention mechanism module is depicted in Figure 2.

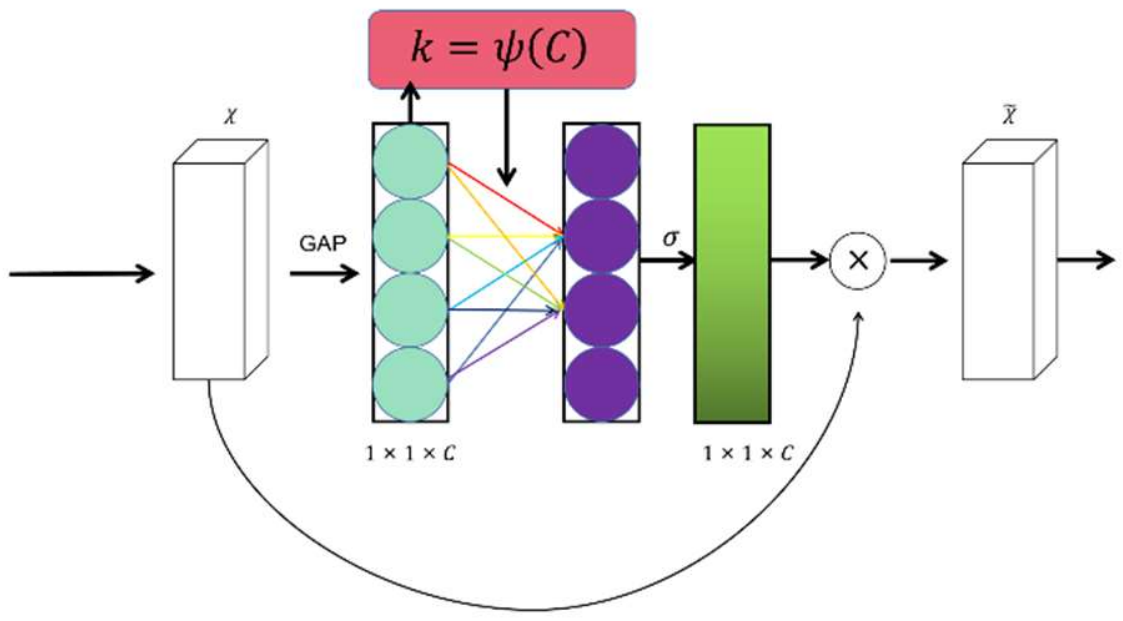

Fig.2 Diagram of adaptive attention module

The ECA attention mechanism is an efficient channel attention mechanism that avoids dimension reduction and enables local cross-channel interactions. Firstly, we apply channel-wise global average pooling to the input feature map, aggregating the convolutional features and obtaining a higher-level representation that summarizes the information across channels, as shown in (1).

$$g(\chi) = \frac{1}{WH}\sum_{i=1,j=1}^{W,H} \chi_{ij} \quad (1)$$

In this context, $\chi$represents the input feature map, and we define y=g($\chi$). Next, we employ a weight matrix Wk to learn channel attention, which effectively captures local cross-channel interaction. The weight matrix Wk is defined as

$$\begin{bmatrix} \omega^{1,1} & \cdots & \omega^{1,k} & 0 & 0 & \cdots & \cdots & 0 \\ 0 & \omega^{2,2} & \cdots & \omega^{2,k+1} & 0 & \cdots & \cdots & 0 \\ \vdots & \vdots & \vdots & \vdots & \ddots & \vdots & \vdots & \vdots \\ 0 & \cdots & 0 & 0 & \cdots & \omega^{C,C-k+1} & \cdots & \omega^{C,C} \end{bmatrix} \quad (2)$$

The weight matrix Wk in this context involves k × C parameters. Channel attention can be learned by

$$\omega_i = \sigma\left(\sum_{j-1}^{k} \omega^j y_i^j\right), y_i^j \in \Omega_i^k \quad (3)$$

In this equation, $\omega_i$ represents the attention weight of the i[th] channel, σ denotes the sigmoid function, $\omega^j$ is the parameter of the convolutional kernel, $y_j^i$ represents the (i+j)[th] channel of the input feature map, and $\Omega_i^k$ denotes the set of the i[th] channel and its k adjacent channels. This design enables all channels to share the same set of learnable parameters. Moreover, this strategy can be effortlessly implemented through the utilization of a fast 1D convolution with a kernel size of k, as shown in the following (4):

$$\omega = \sigma\left(C1D_k(y)\right) \quad (4)$$

C1D represents the one-dimensional convolution operation, which is used in the equation. The size of the kernel in the one-dimensional convolution, denoted by k, determines the coverage range of channel interaction. The value of k is proportional to the channel dimension C. By specifying the channel dimension C, the kernel size k can be adaptively determined, as shown in the following (5):

$$k = \psi(C) = \left|\frac{log_2(C)}{\gamma} + \frac{b}{\gamma}\right|_{odd} \quad (5)$$

In the equation, the term $|t|_{odd}$ denotes the nearest odd number to $t$. In all experiments conducted in this study, we set $\gamma$ and $b$ to 2 and 1, respectively. Finally, the original feature map is element-wise multiplied by the attention weights generated by ECA, resulting in the following (6):

$$\tilde{\chi} = \omega \odot \chi \tag{6}$$

In the equation, $\odot$ represents the Hadamard product, $\tilde{\chi}$ represents the feature map after being weighted by attention, $\omega$ denotes the attention weights, and $\chi$ represents the original feature map. By employing this approach, the ECA attention mechanism can effectively learn and utilize the local cross-channel interactions in the input feature map, providing favorable conditions for further optimization of the model.

In summary, the incorporation of the ECA module significantly improves the network's capability to detect small objects by effectively addressing dimension reduction and facilitating local cross-channel interactions. This attention mechanism plays a crucial role in enhancing the accuracy of detecting small objects in railway construction sites.

### C. *EIoU loss function*

YOLOv5 employs the GIoU loss function[23], which combines the minimum bounding rectangle of the predicted and ground truth boxes. This loss function, compared to the IoU loss, addresses the challenge of predicting accurate distances when the predicted and ground truth boxes are not intersecting. The GIoU loss function is defined as follows, according to (7):

$$GIoU_{Loss} = IoU - \frac{|A_c - U|}{|A_c|} \tag{7}$$

In this context, $A_c$ refers to the minimum enclosing area of the predicted bounding box and the ground truth bounding box. $IoU$ represents the ratio of the intersection to the union between the predicted box and the ground truth box, while $U$ represents the union of the predicted box and the ground truth box.

However, in practical railway construction scenarios, the presence of various equipment and construction materials can often lead to partial occlusion of the targets. This occlusion can significantly affect the adequacy of feature extraction, resulting in potential missed detections and false alarms. During the model training process, it is common for the predicted bounding boxes to primarily capture the unobstructed portions of the targets, neglecting the occluded regions.

During the model training process, it is common for the predicted bounding boxes to only capture the unobstructed portions of the targets, meaning that the predicted boxes are confined within the unoccluded regions of the ground truth boxes. In such cases, the GIoU (Generalized Intersection over Union) metric struggles to accurately determine the positional relationship between the predicted and ground truth boxes. As a result, the loss function remains largely unchanged, leading to slow convergence and reduced detection accuracy of the model.

The following section of this paper will illustrate the limitations of GIoU through a specific example, as shown in Figure 3 and Figure 4. These figures depict a sample image with occluded targets.

In the figures, the blue boxes (smaller boxes) represent the predicted bounding boxes, while the green boxes (larger boxes) represent the ground truth bounding boxes. Figure 3 and Figure 4 correspond to the predicted results before and after fitting, respectively.

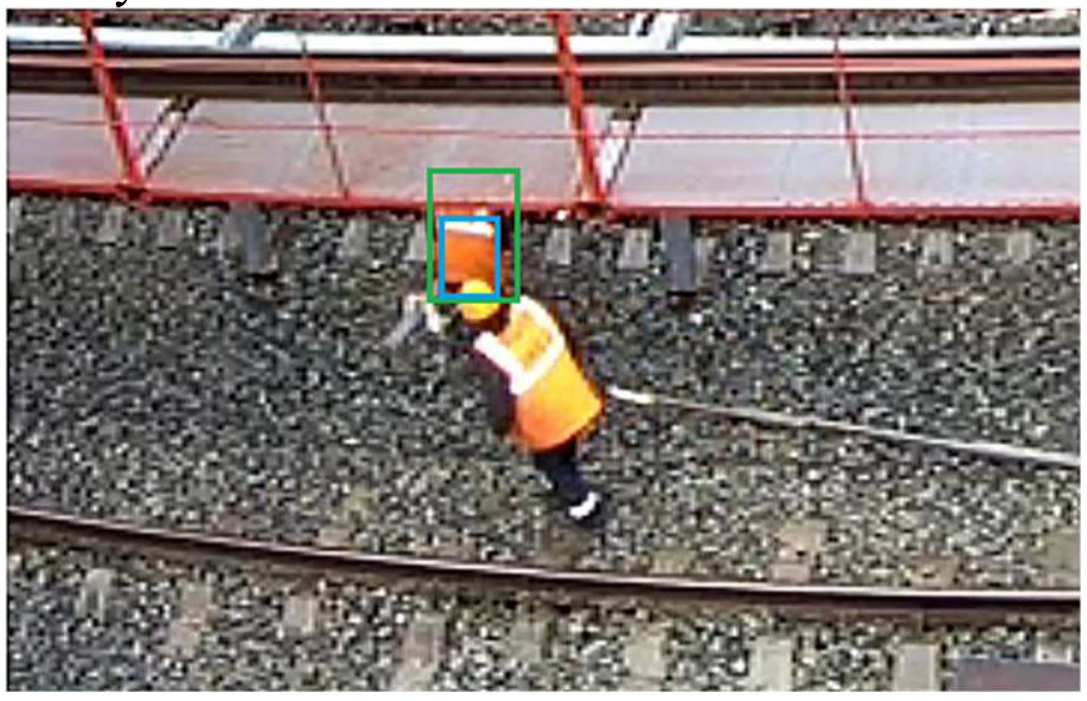

Fig.3 Predicted results before fitting

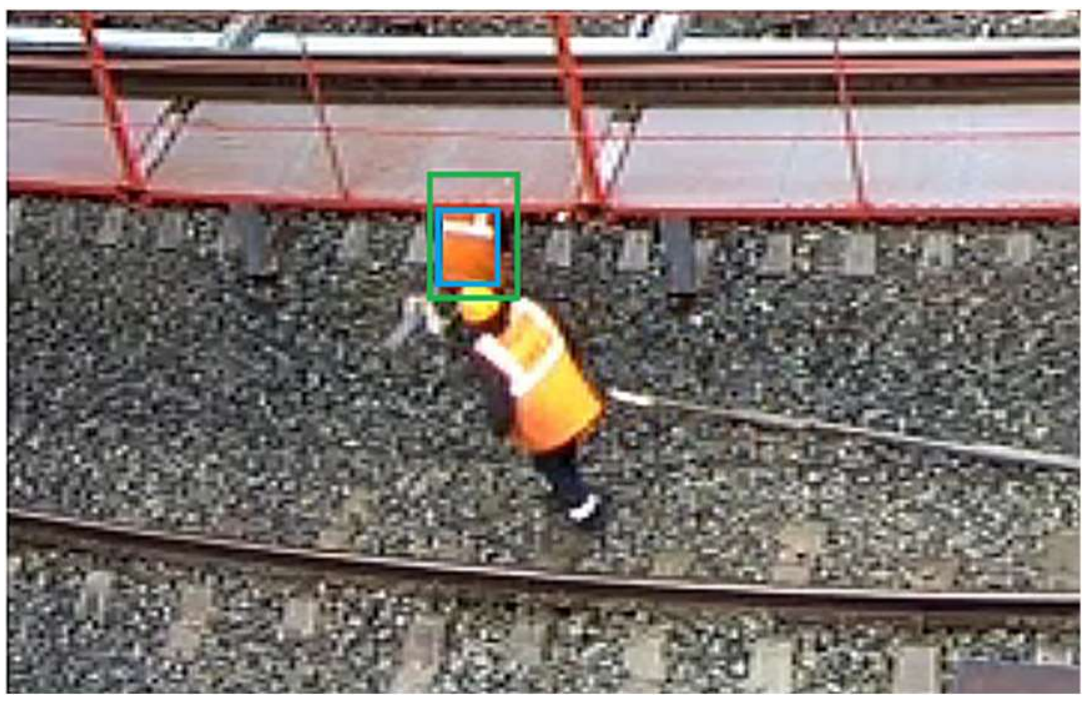

Fig.4 Predicted results after fitting

The following table provides the coordinates of the top-left and bottom-right corners for the boxes shown in the figures:

| Box | Xmin | Ymin | Xmax | Ymax |
|---|---|---|---|---|
| ground truth bounding box | 213 | 83 | 257 | 146 |
| Predicted Box (Before Fitting) | 219 | 107 | 247 | 145 |
| Predicted Box (After Fitting) | 219 | 100 | 247 | 138 |

Table 1 Coordinates of Boxes in Figures

When using GIoU to calculate the loss function before and after fitting, the result is consistently 0.3838. This indicates that the model fitting in this case did not improve accuracy. However, in reality, we can intuitively perceive that the prediction performance after fitting is significantly better than before. Clearly, GIoU is unable to achieve this. Therefore, when training the protective equipment recognition model, it is necessary to introduce a loss function that can determine the positional relationship between the predicted bounding box and the actual box when the predicted box is inside the actual box.

To tackle this issue, we propose the utilization of the EIoU loss function[25]. The EIoU loss function is formulated as shown in (8).

$$EIoU_{Loss} = 1 - IoU + \frac{\rho^2(b, b^{gt})}{c^2} + \frac{\rho^2(\omega, \omega^{gt})}{C_\omega^2} + \frac{\rho^2(h, h^{gt})}{C_h^2} \tag{8}$$

Where $IoU$ represents the ratio of the intersection to the union between the predicted bounding box and the ground truth box. $\rho^2(b, b^{gt})$ is the Euclidean distance between the centers of the predicted box and the ground truth box. $\rho^2(\omega, \omega^{gt})$ is the Euclidean distance between the widths of the predicted box and the ground truth box. $\rho^2(h, h^{gt})$ is the Euclidean distance between the heights of the predicted box and the ground truth box. $c^2$ is the distance between the predicted box and the ground truth box's minimum bounding rectangle diagonal. $C_\omega^2$ represents the closure width between the predicted box and the ground truth box. $C_h^2$ represents the closure height between the predicted box and the ground truth box.

Compared to the GIoU loss function, the EIoU loss function takes into account the distance between the centers of the predicted box and the ground truth box, addressing the issue of the GIoU loss function's inability to determine the positional relationship between the predicted and ground truth boxes when there is an inclusion relationship, i.e., when the ground truth box contains the predicted box. By considering the overlap between the predicted box and the ground truth box, as well as directly computing penalties for width and height, the EIoU loss function improves the convergence speed and regression accuracy of the model.

When using the EIoU loss function to calculate the loss value before and after fitting, the function value increased from 0.0660 before fitting to 0.0895, demonstrating the effectiveness of the model fitting. By utilizing the EIoU loss function, it becomes possible to determine the positional relationship between the predicted and ground truth boxes when there is an inclusion relationship, thereby accelerating the convergence speed and improving the regression accuracy. This approach has been validated through experimentation, confirming the enhanced generalization ability of the model. In practical applications, the improved YOLO-EA model exhibits faster convergence speed and lower loss values, thereby achieving the desired accurate detection results for protective equipment.

## III. RESULTS

### A. Experimental environment

Since this paper focuses on detecting protective equipment in real-world railway construction sites, the experimental hardware used in this study comprised an NVIDIA GeForce RTX 3070 Ti 8G GPU, an Intel i5-13490 CPU with 16GB of RAM, and a 1TB solid-state drive. These hardware specifications are comparable to the mainstream configurations commonly employed in training platforms for target detection algorithms. This choice ensures a more robust evaluation of the effectiveness of the protective equipment detection network model in railway construction scenarios and facilitates an accurate assessment of its practicality.

### B. Experimental Dataset

The dataset used in this study is sourced from the Railroad Worker Detection Dataset on the Kaggle platform[26]. To adapt to the complexity and diversity of railway construction environments, the collected image samples include workers wearing protective equipment in various railway construction scenes and at different times of the day, ensuring that the model possesses sufficient generalization capability. The selected dataset consists of three categories: people, reflective vests, and helmets. The tag "helmets" refers to a type of head protection gear known as a hardhat. After manual counting, the dataset contains 7,884 samples of reflective vests, 6,516 samples of hardhats, and 7,974 samples of people. Among them, there are a total of 1,526 samples where protective equipment is not correctly worn.

In this study, occluded targets are defined as those with an occlusion area less than 35% but greater than or equal to 5%. In the dataset, there are 1,306 occluded samples of reflective vests (16.56% of the total) and 914 occluded samples of hardhats (14.02% of the total), as shown in Table 2. Please refer to Figure 5 for visual representations of occlusion examples. It is worth noting that the training, validation, and test sets were carefully selected to ensure there is no overlap between them. This meticulous approach guarantees the reliability of evaluating the model's performance. Therefore, in this experiment, a total of 4,118 images were used for training, validation, and testing, with 2,290 images used for the training set, 441 images for the validation set, and 491 images for the test set.

| Label | Negative Samples | Occluded Samples | Total Samples |
|---|---|---|---|
| Reflective Vest | | 1306 | 7884 |
| Helmet | 1526 | 914 | 6516 |
| Person | | / | 7974 |

Table 2 Railway site construction dataset

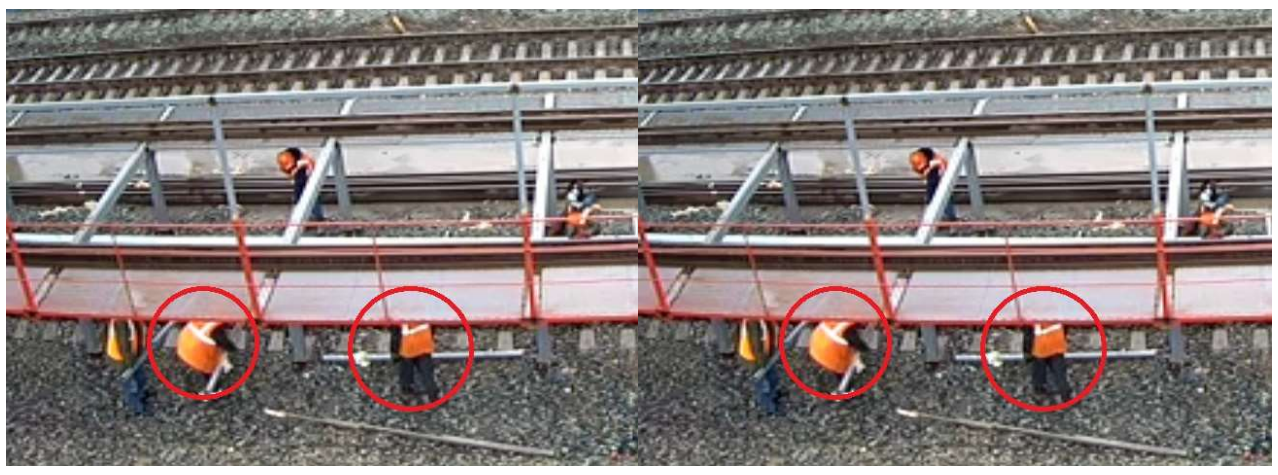

Fig.5 Railway construction site occlusion sample example

### C. Experimental results and analysis

To address the issue of detecting protective equipment worn by construction workers in railway construction sites under natural scenes, this study proposes two improvement measures based on the YOLOv5 architecture and develops the YOLO-EA network model suitable for this scenario. In this section, the impact of these two improvement methods on the model's performance will be investigated through ablation study.

The addition of the attention mechanism module has a significant effect on improving the accuracy of the object detection network. To investigate the impact of the ECA attention mechanism module added in this study on the accuracy of YOLOv5 in protective equipment detection, we conducted ablation study with two groups of neural network models. One group used the original YOLOv5 model, while the other group used the YOLOv5 model with the additional ECA attention mechanism module. The remaining settings were kept the same as YOLOv5. The results of the ablation study are shown in Table 3. With the inclusion of the ECA attention network module, the precision reached 98.2%, with an mAP50-95 of 0.671, which is a 2.4% improvement compared to the original YOLOv5 model. Furthermore, when the loss function was changed to EIoU, the precision improved by nearly 1%.

| ECA | EIOU | mAP50-95 | precision /% | recall/% | Epochs |
|---|---|---|---|---|---|
| | | 0.647 | 96.4 | 94.2 | |
| √ | | 0.671 | 98.2 | 95.5 | 200 |
| | √ | 0.642 | 97.2 | 94.8 | |
| √ | √ | 0.692 | 98.9 | 94.7 | |

Table 3 Model ablation study

The experiments demonstrate that the incorporation of the ECA attention mechanism module in the YOLOv5 network architecture helps the model extract information from small object categories, thereby effectively improving the precision of the YOLOv5 network model. Additionally, in YOLOv5, the model can accurately determine the positional relationship when there is an inclusion between the predicted box and the ground-truth box. This approach accelerates the convergence speed and enhances the regression precision. This approach accelerates the convergence speed and enhances the regression precision.

*D. Analysis of ECA Attention Mechanism on Small Object Detection*

To conduct a detailed analysis of the effectiveness of the ECA attention mechanism in detecting small objects, this study selected an image containing small objects as a case study (as depicted in Figure 6). The analysis compared two models: YOLOv5 and YOLO-ECA. The former represents the YOLOv5 model without the integration of the ECA attention mechanism, while the latter incorporates the ECA attention mechanism into the YOLOv5 model.

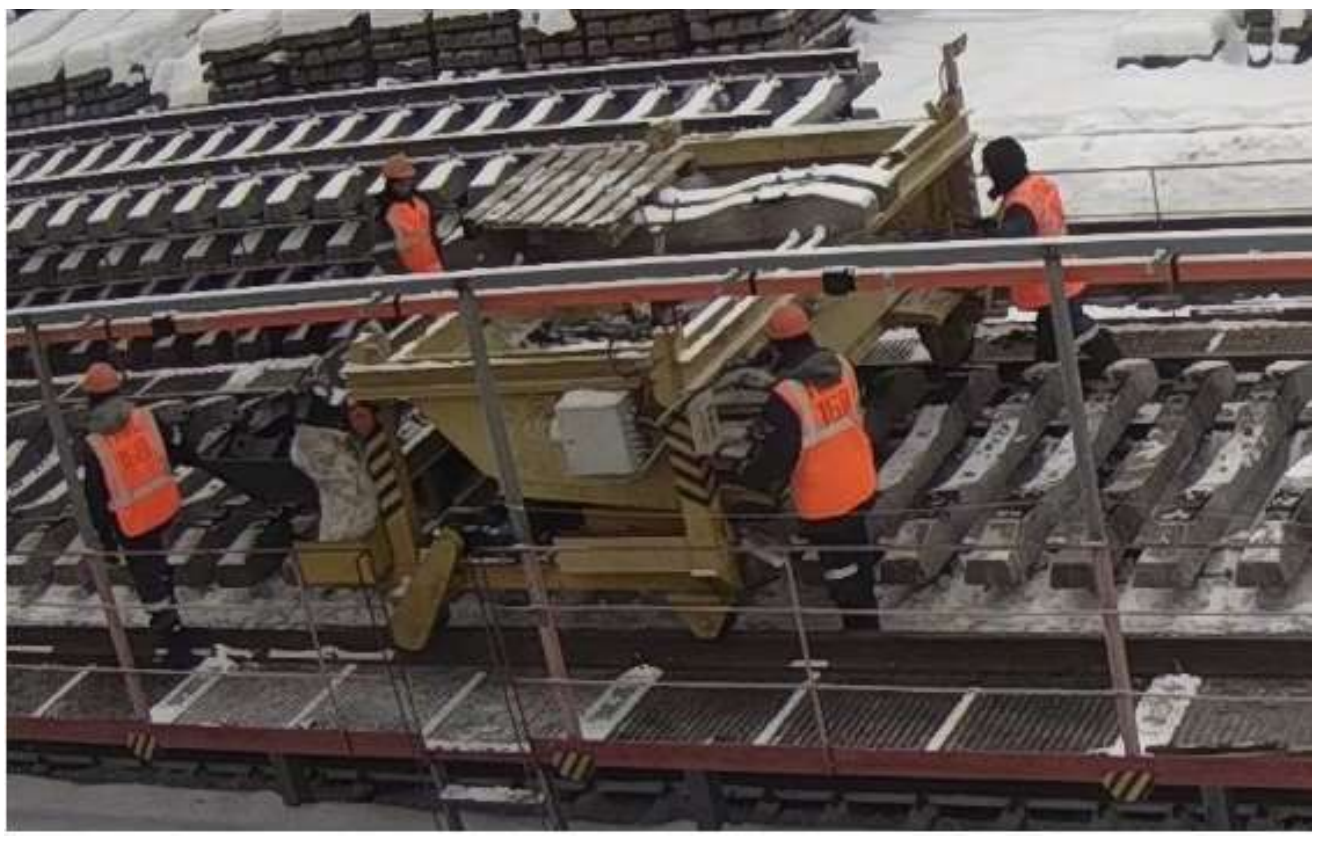

Fig.6 small target sample

The selected sample image, shown in Figure 6, displays four workers near a road roller, each assigned with a corresponding number for identification. Specifically, the worker without a hardhat is labeled as worker number one, while the remaining three workers are labeled as workers number two, three, and four in a clockwise direction. It should be noted that workers number one and four are relatively smaller in size compared to workers number two and three.

To provide a more intuitive analysis of the effectiveness of the ECA attention mechanism, we utilized smooth-type thermal diagram for observation. Smooth-type thermal diagram were employed to visually represent the attention focus and highlight the regions of interest in the image.

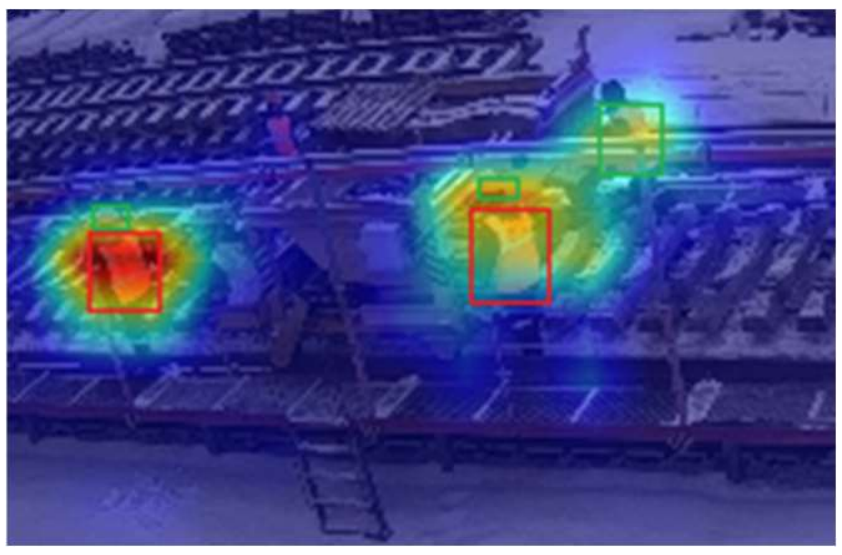

Fig.7 yolov5

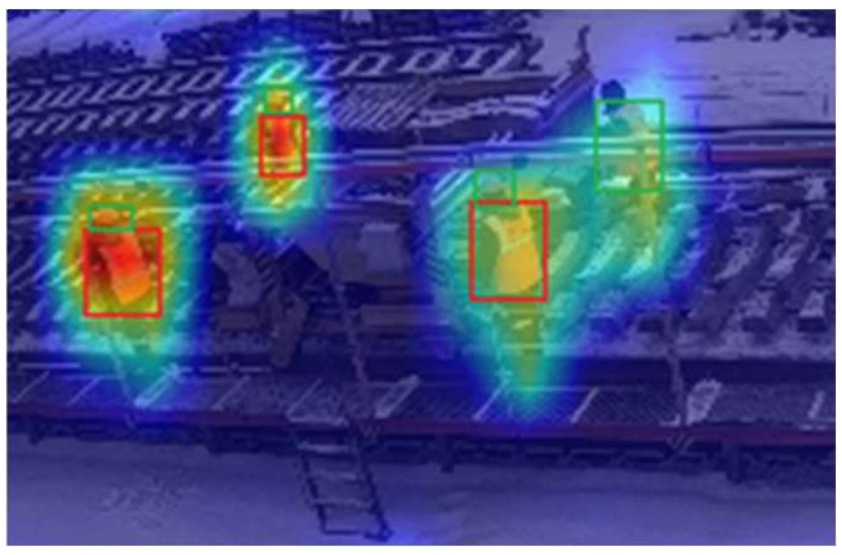

Fig.8 YOLO-ECA

In Figure 7, we present the smooth-type thermal diagrams generated by the YOLOv5 model. It is evident that the focus of the YOLOv5 smooth-type thermal diagrams is primarily on workers one, two, and three, while the attention given to worker four is relatively limited. Moreover, for larger targets, such as worker three, the YOLOv5 smooth-type thermal diagrams exhibit the highest level of concentration, represented by the color red.

In contrast, Figure 8 showcases the smooth-type thermal diagrams generated by the YOLO-ECA model, which incorporates the ECA attention mechanism. It is evident that the YOLO-ECA smooth-type thermal diagrams not only concentrate on the regions of interest identified by YOLOv5 (i.e., workers one, two, and three) but also exhibit additional focus on worker four, a smaller target. This observation strongly indicates the efficacy of the ECA attention mechanism in recognizing small objects.

In conclusion, the comparison of the smooth-type thermal diagrams generated by the YOLOv5 and YOLO-ECA models leads us to the following conclusion: the introduction of the ECA attention mechanism improves the detection of small objects, enabling the model to focus more accurately on these targets. This finding holds significant implications for the detection of small objects in the context of railway site construction scenarios.

## E. *EIOU's effect analysis of blocking target*

In order to thoroughly analyze the application effectiveness of the EIOU loss function for occluded object samples in object detection tasks, we have selected an image with occluded objects (as shown in Figure 9) and conducted detection analysis using two models. The selected models are YOLOv5, which utilizes the conventional GIOU loss function, and YOLO-EIOU, which utilizes the EIOU loss function.

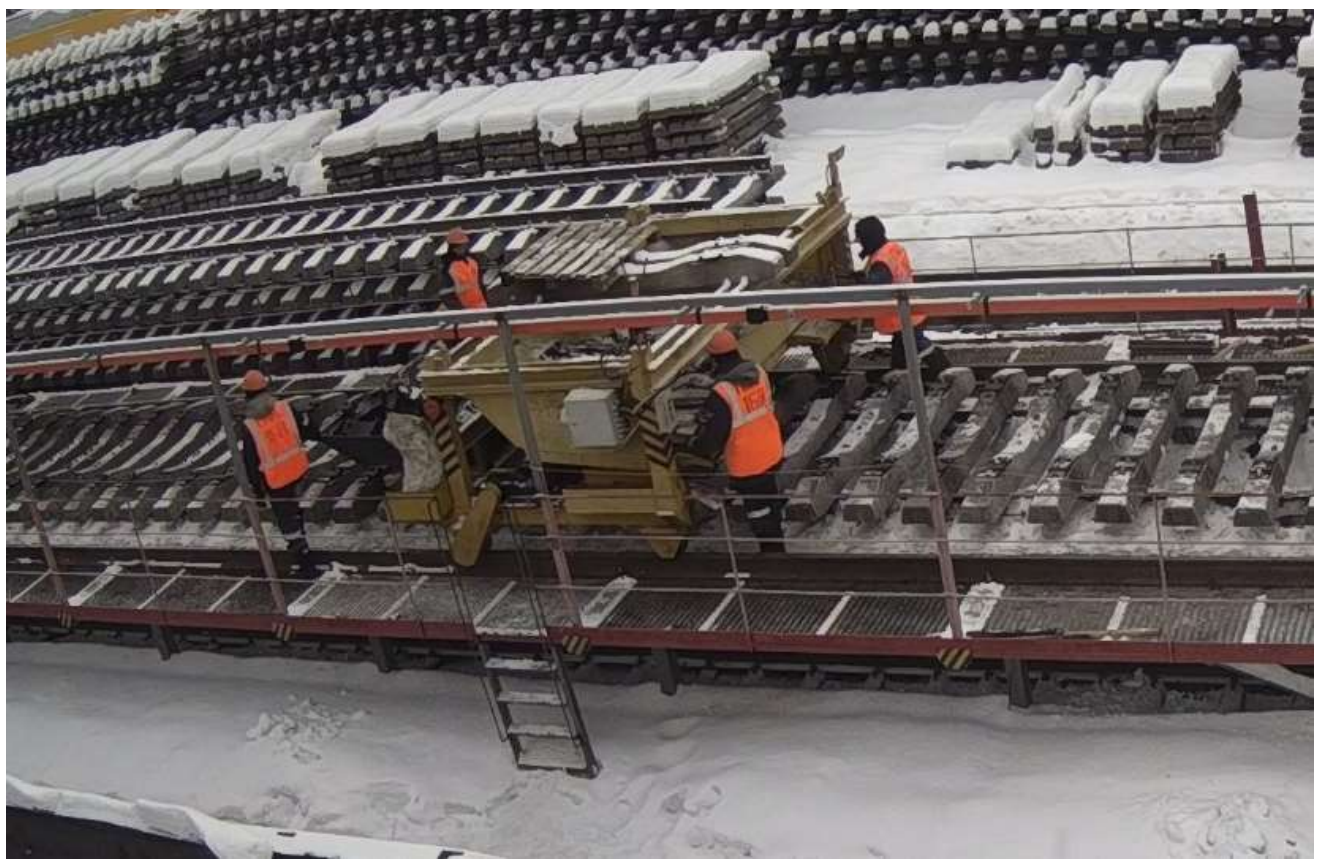

Fig.9 Target occlusion sample

For ease of analysis, we have assigned numbers to the four workers next to the road roller in the image. The worker without a hardhat is labeled as Worker 1, while the remaining three workers are labeled in a clockwise direction as Worker 2, Worker 3, and Worker 4. It is important to note that the reflective vests of Worker 1 and Worker 4 are partially occluded. Worker 1's vest is obstructed by the railing at the site, while Worker 4's vest is obscured by the road roller. On the other hand, the reflective vests of Worker 2 and Worker 3 are not occluded.

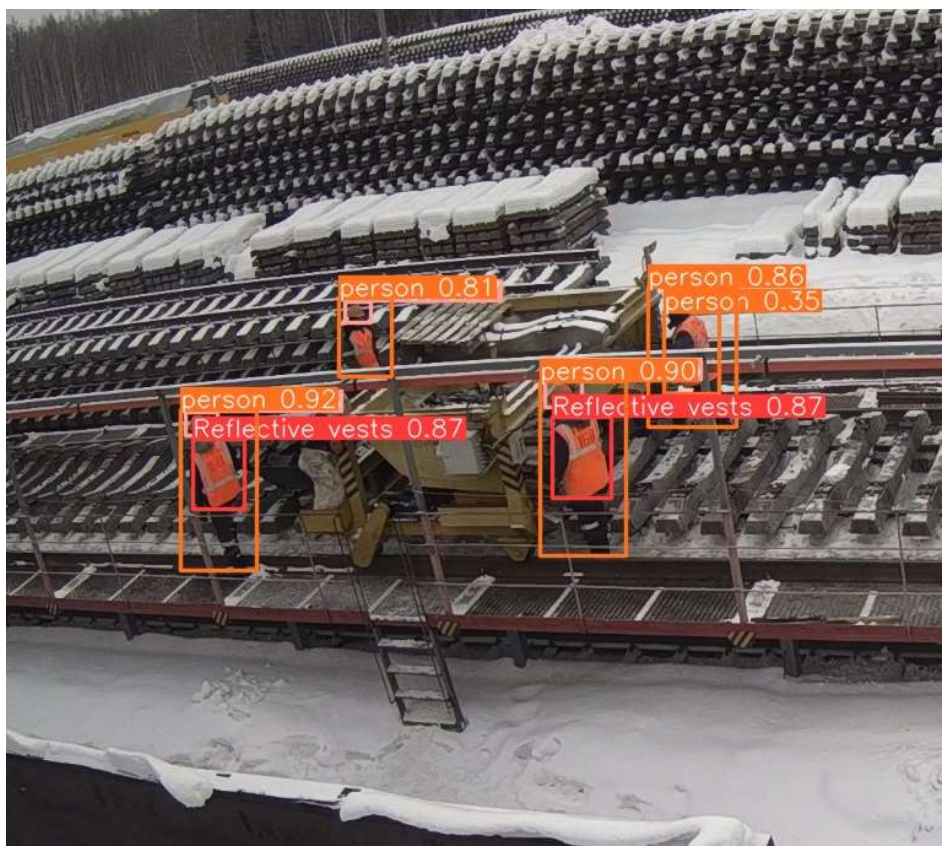

Fig.10 YOLOv5

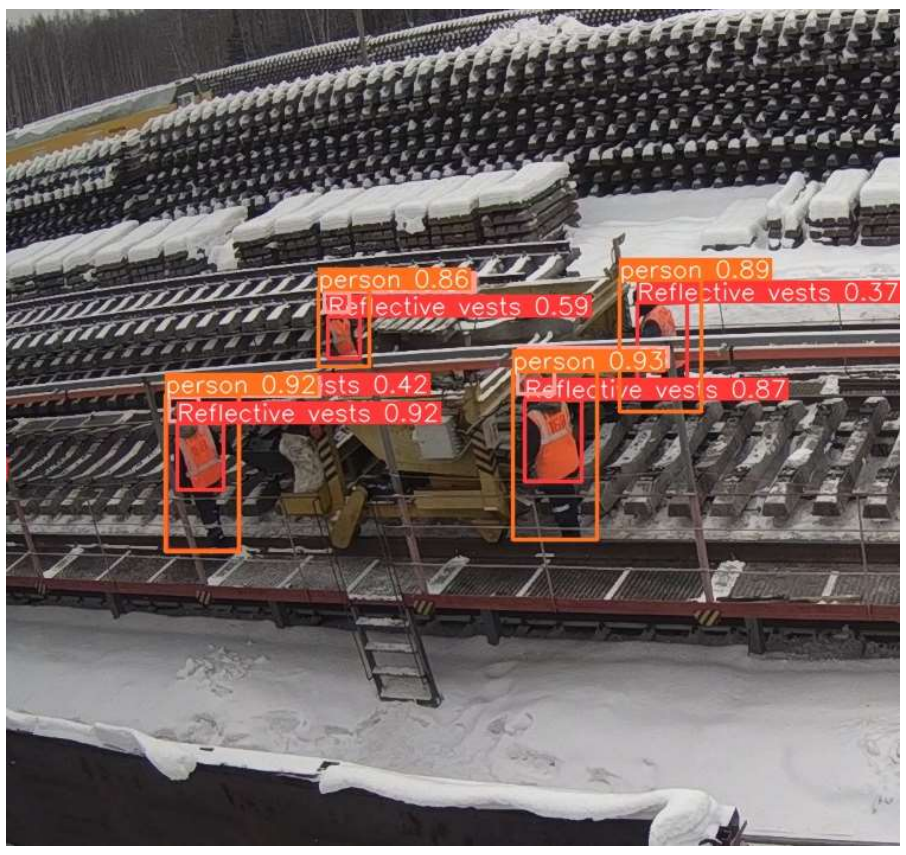

Fig.11 YOLO-EIOU

According to Figure 10, the YOLOv5 model trained with the GIOU loss function is able to accurately detect the complete set of protective equipment for Worker 2 and Worker 3. However, it struggles to accurately predict the reflective vests of occluded Worker 1 and Worker 4. This can be attributed to the limited performance of traditional loss functions in handling occlusion and the insufficient fitting accuracy of the model.

However, the analysis results in Figure 11 indicate that the YOLO model trained with the EIOU loss function demonstrates a significant advantage in handling occlusion issues, particularly for Worker 1 and Worker 4. It successfully predicts the bounding boxes of the occluded objects. This can be attributed to the EIOU loss function's incorporation of overlap between predicted and ground-truth boxes, as well as direct calculations of width and height, resulting in improved regression accuracy and enhanced robustness of the model when dealing with occlusion.

The results of this quantitative comparison clearly demonstrate that the EIOU loss function exhibits significant advantages over traditional loss functions when it comes to handling occlusion issues. It leads to a substantial improvement in the model's fitting accuracy.

## F. *Training process analysis*

This study evaluates the performance of the detection models using loss function curves and mAP50-95 curves. The loss function curves depict the convergence speed and convergence level of the network models during training. Figure 12 illustrates the loss function curves of the original YOLOv5 model and the improved YOLO-EA model.

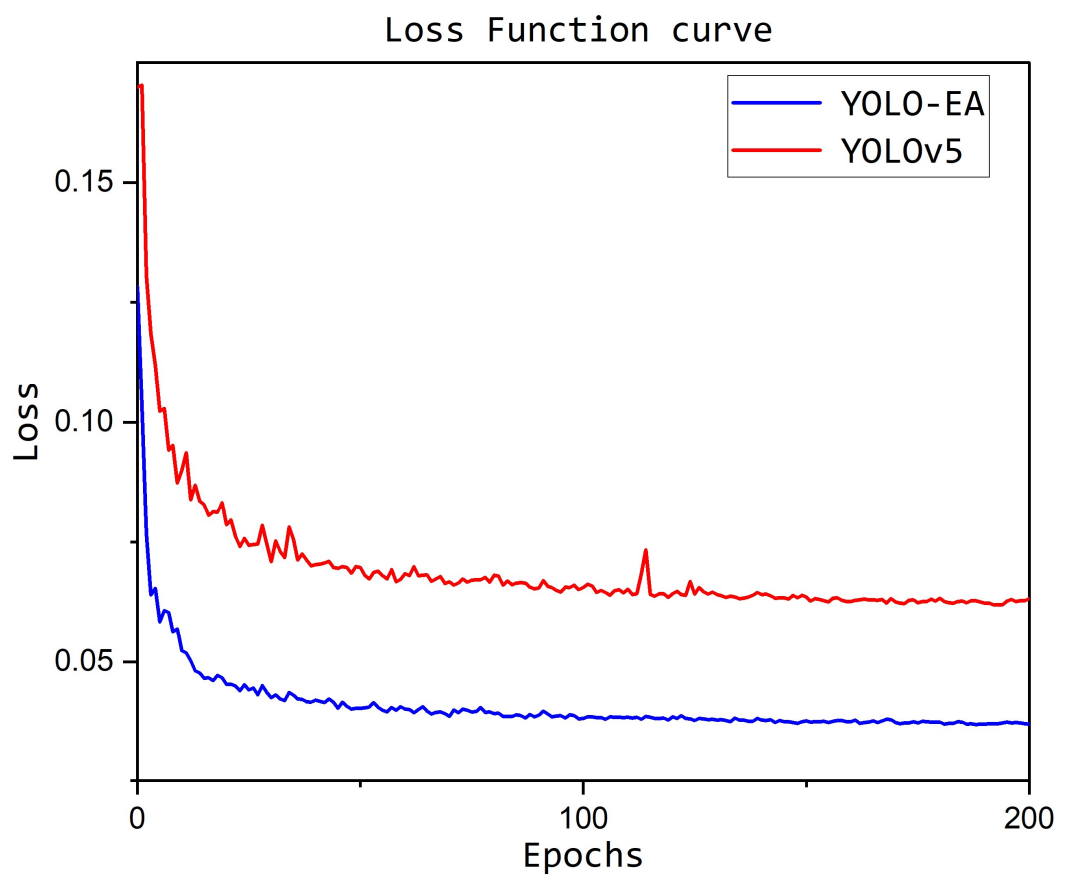

Fig.12 Loss Function curve

As shown in the above figure, during the training process of 200 epochs, the improved YOLO-EA model exhibits a faster convergence speed and lower loss values compared to the original YOLOv5 model. Upon completion of training, the improved YOLO-EA model achieves a 71.04% lower loss value than the original YOLOv5 model, demonstrating its superior convergence performance.

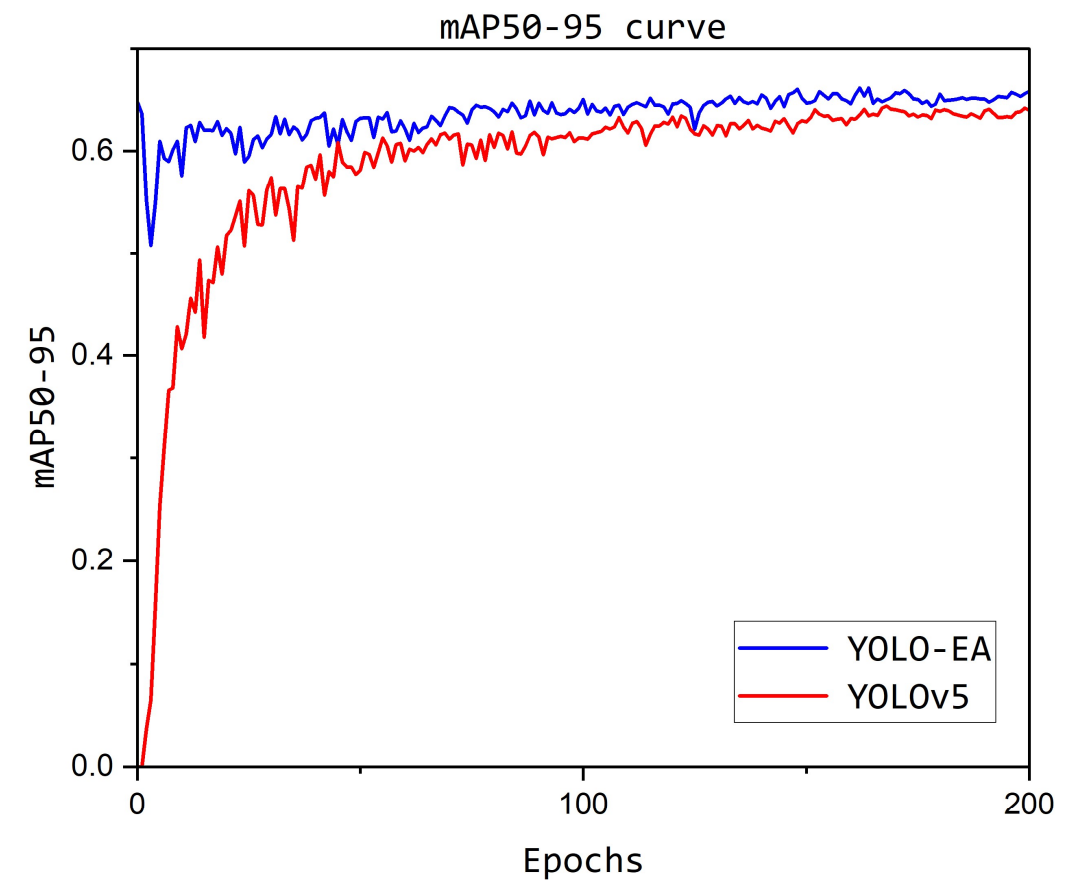

Fig.13 mAP50-95 curve

We evaluated the detection performance of the models using mAP50-95, where a higher value indicates better detection performance. As shown in Figure 13, after 200 epochs, the improved YOLO-EA model achieves a stable mAP50-95 value of 0.658, with a peak value of 0.661 during the training process. This indicates that the improved YOLO-EA model demonstrates good performance in protective equipment detection, achieving the desired accurate recognition results as expected.

## *G. Target detection performance in various categories*

In this section, we provide a detailed analysis of the accuracy of our object detection model across different categories. In the ablation study, we have already explored which components in the model architecture are crucial for improving accuracy. By further analyzing the accuracy across different categories, we can gain insights into the model's performance and limitations in different object classes, thus providing targeted guidance for optimization. Table 4 presents the performance evaluation of our YOLO-EA model in various categories:

| Object Category | Precision /% | Recall /% | mAP50-95 |
|---|---|---|---|
| Reflective vest | 98.5 | 92.6 | 0.686 |
| Helmet | 98 | 94.6 | 0.635 |
| Person | 98.5 | 98.6 | 0.768 |

Table 4 YOLO-EA model performance evaluation in various categories

From the table above, we can observe that YOLO-EA performs the best in the Person category, with an mAP50-95 of 0.768. However, in the case of protective equipment, the metrics are relatively lower, with values of 0.686 and 0.635 for Reflective vest and Helmet, respectively. This difference can be attributed to the smaller and more complex shapes of protective equipment compared to the relatively larger and more abundant human objects in the dataset. The model's performance is influenced by the size and complexity of the objects, with better results achieved for larger and more prevalent categories such as humans.

In this model, we focus on detecting humans and protective equipment with the aim of implementing an Intersection over IoU calculation strategy to improve the detection accuracy of protective equipment. For example, in actual railway construction site scenarios, hardhats and reflective vests may be hung nearby instead of being worn, and excluding humans from the detection targets could significantly increase the model's false positive rate. By conducting in-depth analysis of the detection performance for each category, we can optimize the IoU threshold specifically for certain categories to achieve more accurate object detection. By adjusting the IoU threshold, we expect to reduce false positives and false negatives for protective equipment, thereby improving overall detection performance. This strategy holds promise in providing strong support for further research and offering more accurate and reliable solutions for the detection of protective equipment in practical railway construction site applications.

## *H. Model performance comparison*

In order to validate the effectiveness of the proposed YOLO-EA network architecture for detecting and recognizing the wearing of protective equipment in railway construction site scenarios, this paper compares the performance of the model with models from the Kaggle platform, YOLOv5x, and YOLOv5 with the addition of the SPD attention mechanism[27]. The evaluation is conducted using the dataset designed specifically for natural railway construction site environments, as shown in Table 5. All models are uniformly trained using 640x640 images as inputs.

| Model | fps | mAP | Precision/% | Recall/% |
|---|---|---|---|---|
| YOLOv5 from Kaggle[28] | 79.371 | 0.559 | 93.7 | 91.8 |
| YOLOv5x | 71.211 | 0.647 | 96.4 | 94.2 |
| YOLO-SPD | 77.744 | 0.601 | 97.5 | 94.0 |
| YOLO-EIOU | 70.922 | 0.642 | 97.2 | 94.8 |
| YOLO-ECA | 70.844 | 0.671 | 98.3 | 95.3 |
| YOLO-EA | 70.774 | 0.692 | 98.9 | 94.7 |

Table 5 Comparison of protective equipment recognition effects of different models

According to the results in Table 5, we observed that the introduction of the ECA attention mechanism module and the use of the EIoU loss function effectively improved the accuracy of the network model, achieving a precision of 98.9%, surpassing other detection algorithms. Furthermore, the model achieved 70.774 fps, meeting the requirements for real-time detection. Compared to other network architectures, the proposed model in this paper demonstrates higher recognition accuracy. The recognition results of different models are shown in Figure 14, where (a), (b), (c), and (d) represent the recognition results of accurate wearing of protective equipment by the construction workers using the object detection algorithm. From Figure (a), it can be observed that the algorithm proposed in this paper outperforms mainstream object detection algorithms in terms of protective equipment detection.

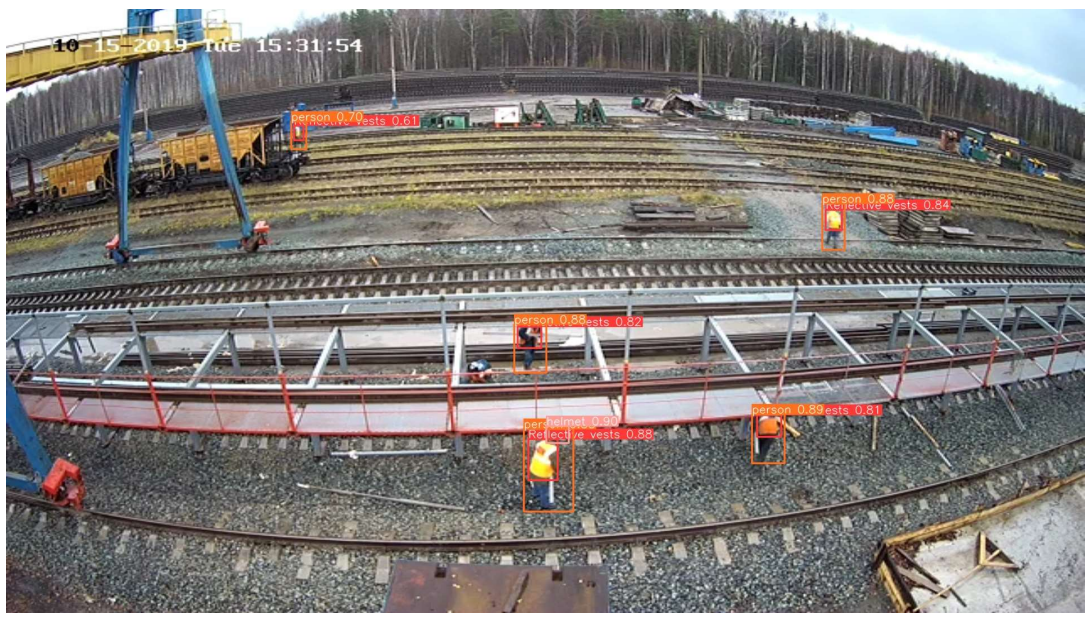

Fig.a YOLO-EA

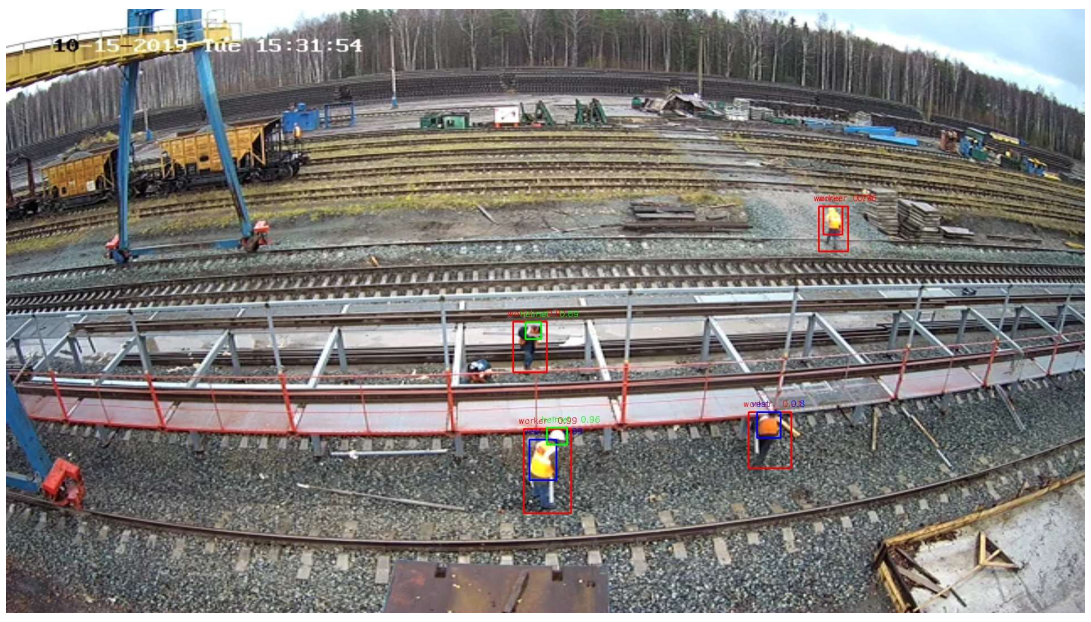

Fig.b YOLO from Kaggle

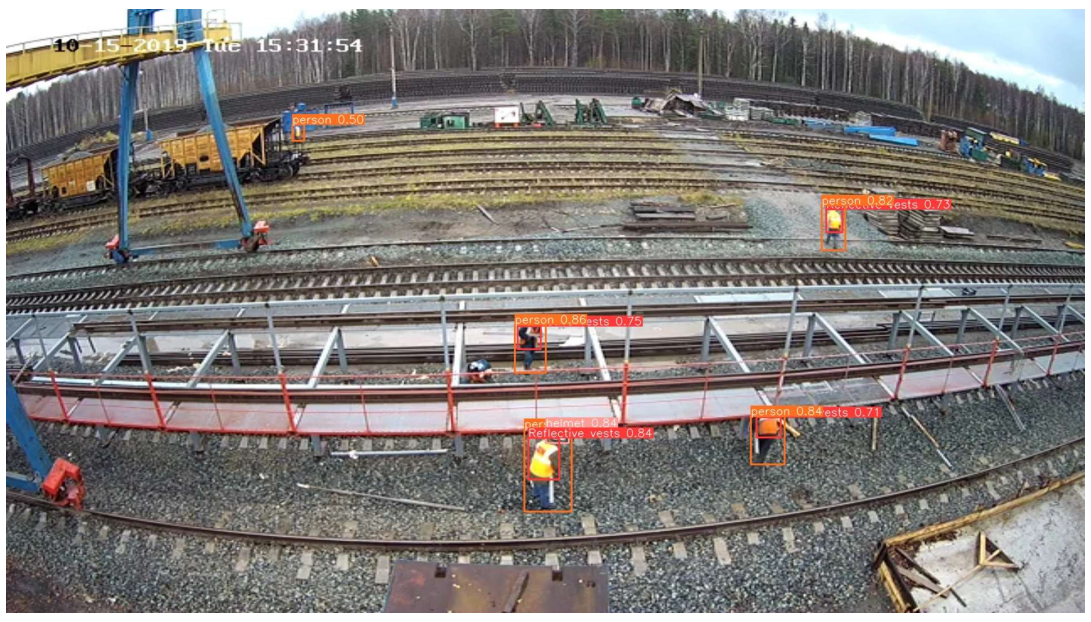

Fig.c YOLO-SPD

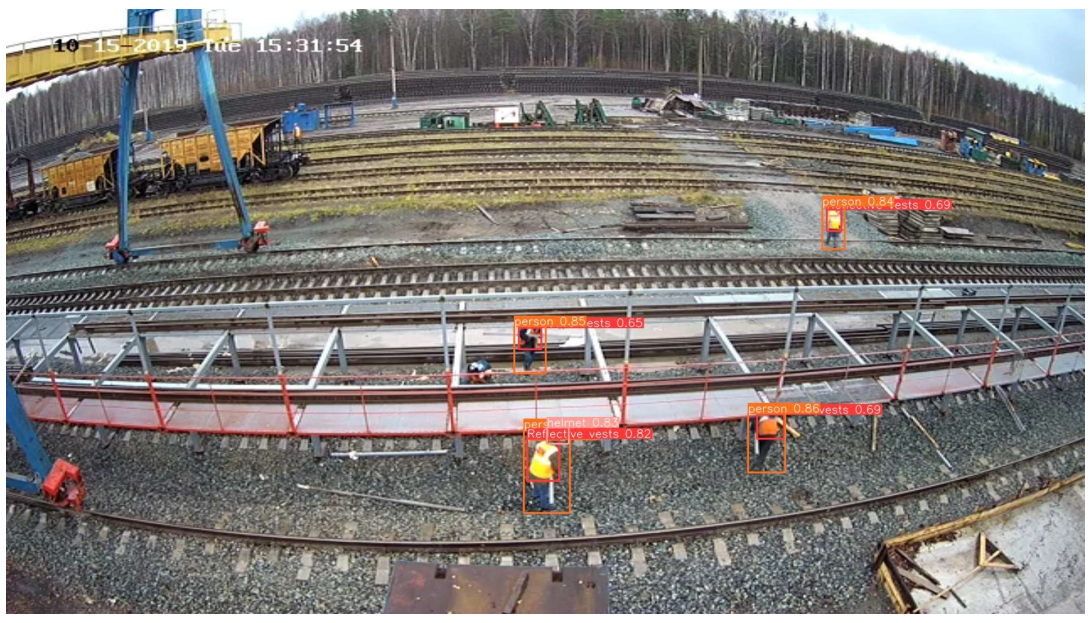

Fig.d YOLOv5

Fig.14 Comparison of hardhats recognition detection effects of different models

## I. *Experimental results analysis*

To validate the recognition performance of the proposed YOLO-EA model in detecting the wearing of protective equipment in natural environments, we conducted tests using a dataset from a railway construction site. Figure 15 shows the recognition results of the YOLO-EA model in railway construction scenarios with occluded targets. In the figure, we can observe that the YOLO-EA model demonstrates robustness in recognizing protective equipment even in the presence of target occlusion.

Figure 16 shows the recognition results of YOLOv5 on a subset of the test dataset, while Figure 17 shows the recognition results of YOLO-EA. It is evident that the addition of the ECA attention mechanism module proposed in this paper and the adoption of the EIOU loss function effectively reduce the false negatives and false positives in detecting small hardhats and reflective vests, thus improving the accuracy of protective equipment detection.

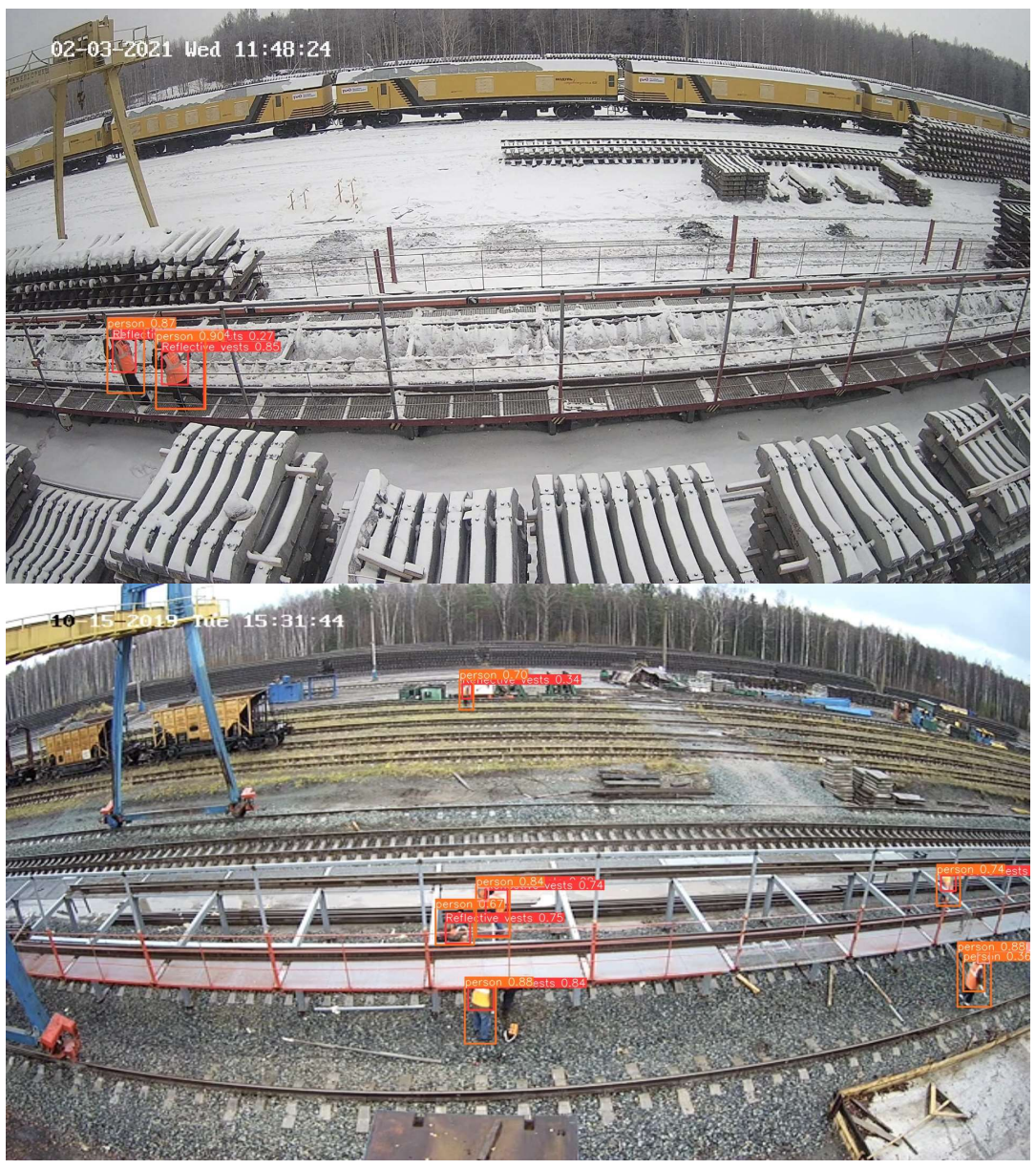

Fig.15 Recognition results of YOLO-EA under object occlusion problem

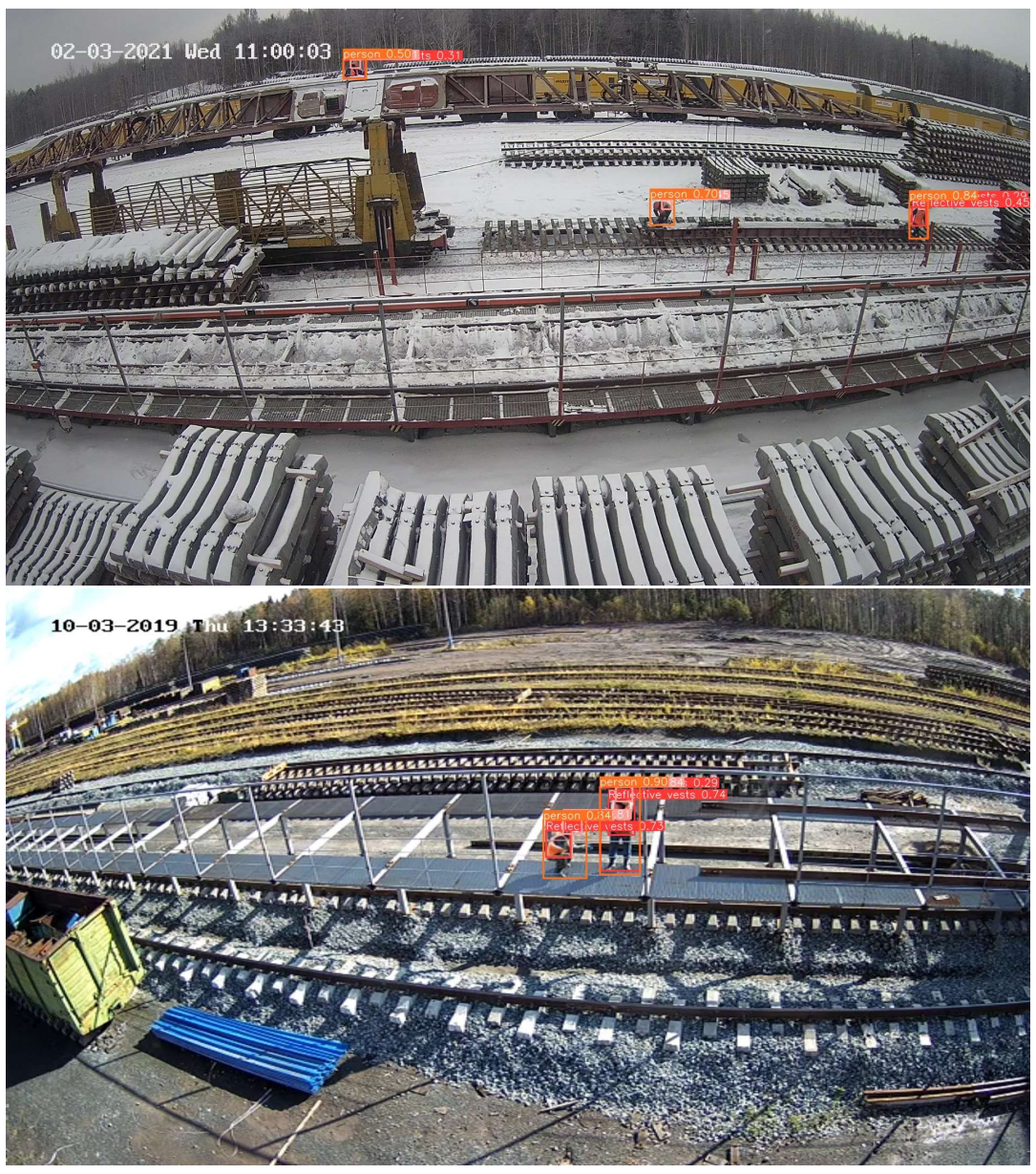

Fig.16 YOLOv5 recognition results

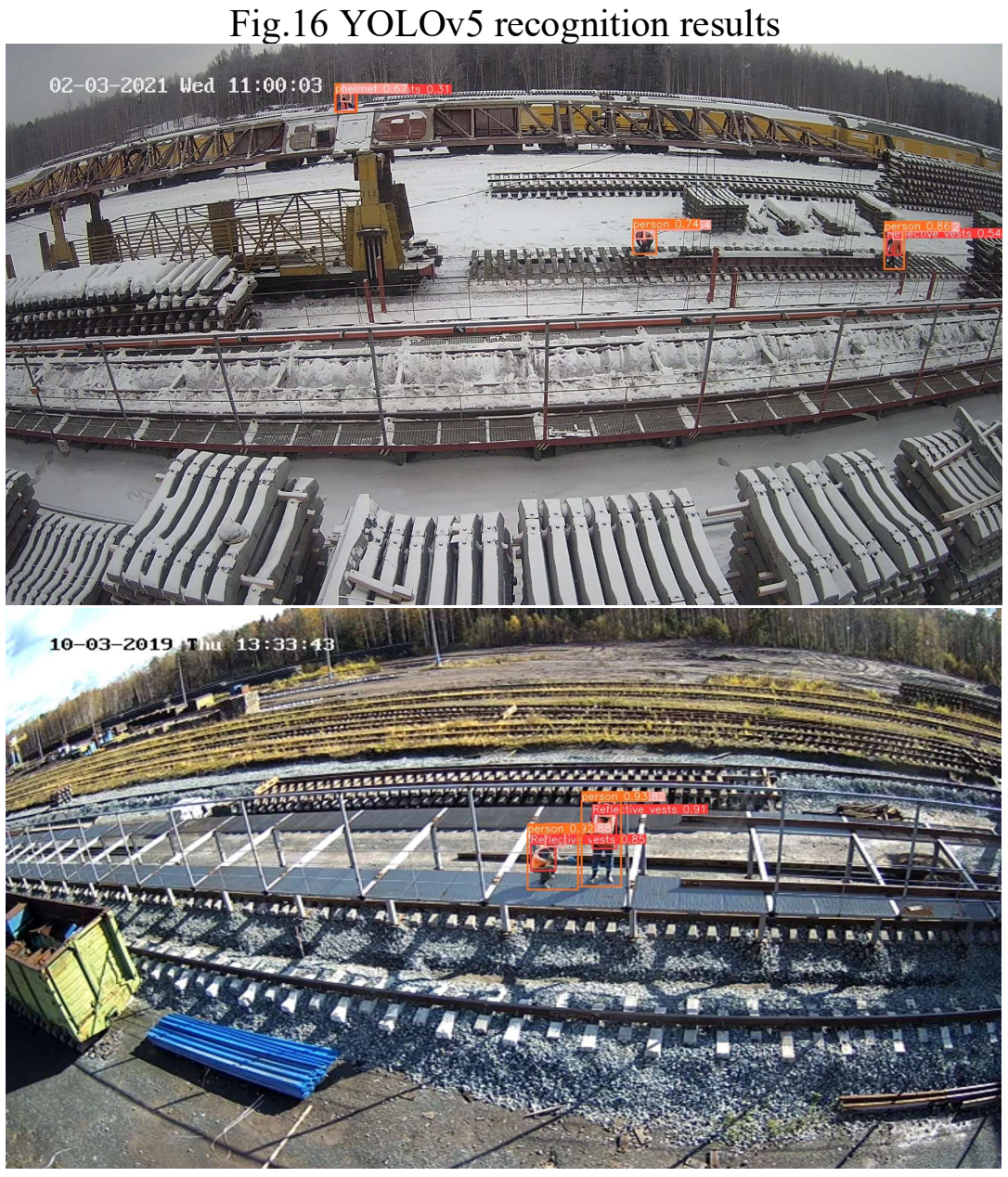

Fig.17 YOLO-EA recognition results

## IV. Results

To address the challenges of recognizing small target objects in railway construction environments and handling target occlusion, this study proposes a neural network model named YOLO-EA based on the YOLO network architecture. The YOLO-EA model incorporates two key innovations. Firstly, it introduces the ECA channel attention mechanism module based on one-dimensional convolutional kernels into the backbone of YOLOv5. This enhancement improves the model's ability to recognize small target objects. Secondly, the EIOU loss function is employed to replace the original GIOU loss function. By considering the aspect ratio and improving upon GIOU, the model's handling of target occlusion is enhanced, leading to faster convergence and improved regression accuracy.

## V. Conclusions

The results of the ablation study demonstrate that the YOLO-EA model outperforms the YOLOv5 model in terms of accuracy and real-time performance in practical application scenarios. Therefore, it can serve as a viable solution to overcome the limitations of manual detection in traditional construction site environments. However, it is important to optimize the model to reduce its parameter size, considering that the algorithm's parameter count remains unchanged. This optimization is necessary to embed the model into smaller IoT devices. In future work, we plan to deploy the YOLO-EA network model in real construction environments using embedded devices. Additionally, we aim to incorporate IoU calculations for both personnel and protective equipment at the output end to identify individuals wearing hardhats. Furthermore, we will explore methods for multi-angle detection of protective equipment wearing behavior to further enhance the model's recognition accuracy.